\documentclass{article} 
\usepackage{iclr2021_conference,times}

\usepackage{graphicx}
\graphicspath{ {./images/} }

\usepackage{subcaption}

\usepackage{amsmath,amsfonts,bm}

\def\eqref#1{equation~\ref{#1}}

\def\1{\bm{1}}

\DeclareMathAlphabet{\mathsfit}{\encodingdefault}{\sfdefault}{m}{sl}
\SetMathAlphabet{\mathsfit}{bold}{\encodingdefault}{\sfdefault}{bx}{n}

\usepackage{hyperref}
\usepackage{url}
\usepackage[ruled,vlined]{algorithm2e}
\usepackage{comment}

\SetCommentSty{mycommfont}

\newcommand{\ctext}[1]{\raise0.2ex\hbox{\textcircled{\scriptsize{#1}}}}

\title{Out-of-core training \\
for extremely large-scale neural networks with adaptive window-based scheduling}

\author{Akio Hayakawa \\
R\&D Center \\
Sony Corporation \\
Osaki, Tokyo 141-8610, Japan \\
\texttt{Akio.Hayakawa@sony.com} \\
\And
Takuya Narihira \\
R\&D Center \\
Sony Corporation \\
Osaki, Tokyo 141-8610, Japan \\
\texttt{Takuya.Narihira@sony.com}
}

\iclrfinalcopy 
\begin{document}



\maketitle
\begin{abstract}
While large neural networks demonstrate higher performance in various tasks, training large networks is difficult due to limitations on GPU memory size. 
We propose a novel out-of-core algorithm that enables faster training of extremely large-scale neural networks with sizes larger than allotted GPU memory. 
Under a given memory budget constraint, our scheduling algorithm locally adapts the timing of memory transfers according to memory usage of each function, which improves overlap between computation and memory transfers. 
Additionally, we apply virtual addressing technique, commonly performed in OS, to training of neural networks with out-of-core execution, which drastically reduces the amount of memory fragmentation caused by frequent memory transfers. With our proposed algorithm, we successfully train ResNet-50 with 1440 batch-size with keeping training speed at 55\%, which is 7.5x larger than the upper bound of physical memory. It also outperforms a previous state-of-the-art substantially, i.e. it trains a 1.55x larger network than state-of-the-art with faster execution. Moreover, we experimentally show that our approach is also scalable for various types of networks.

\end{abstract}

\section{Introduction}

Deep Neural Networks (DNNs) achieve outstanding results in various tasks. In particular, it has been demonstrated that larger neural networks outperform smaller ones. For example, on image classification tasks, \cite{he2016identity} shows ResNet with 1k-layers achieves accuracy improvement by 2\% over ResNet-110 without any changes except the number of layers. Likewise, larger models achieve better performances on natural language processing \citep{devlin2018bert, brown2020language} and image generation \citep{wang2018pix2pixHD, brock2018biggan}. Supported by these evidences, making model larger is one promising way to improve the model performance and realize brand-new systems in the deep learning research and development.

Despite the high demand for large models, the GPU memory size is limited. For instance, NVIDIA A100, one of the latest GPU devices, has only 40GB as its memory. Such limitation on memory inevitably places an upper bound on the scope with which deep learning researchers and developers design architectures of neural networks. It also limits the capacity of neural networks to perform better on existing tasks or deploy richer amount of data that are not tractable on current GPU limitations, such as 4K videos, 3D contents, and so on.

One possible way to address the limitations on GPU memory size is ``out-of-core execution". This method utilizes CPU memory as a temporary cache for the GPU computation. Since neural networks, especially feed-forward networks, can be executed layer by layer sequentially, we can transfer data from GPU to CPU memory when the variables are not necessary at the current computation. In fact, the CPU memory size is much larger than GPU memory, e.g., larger than 1TB. Thus, using CPU memory as a cache for GPU memory, we can virtually extend the size of GPU memory, as if it has memory larger than 1TB. 

As a naive strategy to realize out-of-core execution, we can transfer memory between GPU and CPU before and after every layer execution. While this approach can execute the maximum size of model on limited memory budget, this approach puts GPU computation on hold at every layer until the end of corresponding memory transfers. On the other hand, if we place too many variables on GPU to accelerate computation, we can execute only models with limited size. Therefore, it is necessary to find a better memory transfer algorithm that enables execution of larger models without sacrificing computational time. 

Many existing works attempted to achieve a faster execution with models larger than allotted GPU memory in the setting with out-of-core execution  \citep{Rhu2016vdnn,Jin2018LayRub,wang2018superneuron}. Recently, \cite{IBM2019} formulated out-of-core execution as inserting memory transfer operations in the graph on TensorFlow \citep{tensorflow2015-whitepaper}. This method can be applied to arbitrary architecture of neural network, and they successfully enable a faster execution with larger model than previous works. However, they only focus on the fixed distance represented by edges on the graph and do not consider memory usage of each function and variable.
It thus results in significant overhead, especially as the model size grows.
Moreover, to the best of our knowledge, none of the previous methods tackled a memory fragmentation problem caused by frequent memory transfers. The memory fragmentation wastes the limited memory budget on GPU and limits the scalability of model size.

In this paper, we introduce a novel memory transfer scheduling algorithm for the training of extremely large-scale neural networks. Our key contributions are summarized as follows:
\begin{itemize}
  \item We propose a novel cost-aware memory swapping scheduling algorithm that is simple yet can find a faster schedule. Our model is the first to schedule memory transfers for training neural networks with locally adaptive distance by modeling the memory usage, which can find a faster schedule than existing methods.
  \item We introduce a novel memory allocation strategy to reduce memory fragmentation caused by memory swapping. Based on a virtual addressing technique that is mainly used in OS, our memory allocation strategy suppresses memory fragmentation and improve the trainable model size.
  \item Through experiments, we validate our proposed method performs well in terms of both training time and model size. On ResNet50, we show our algorithm clearly outperforms a previous work. Additionally, we evaluate our method on the various networks for image recognition, semantic segmentation, and image generation to show our approach can be applied to various tasks. 
\end{itemize}
 
\begin{figure}[t]
\centering
\includegraphics[width=0.9\textwidth]{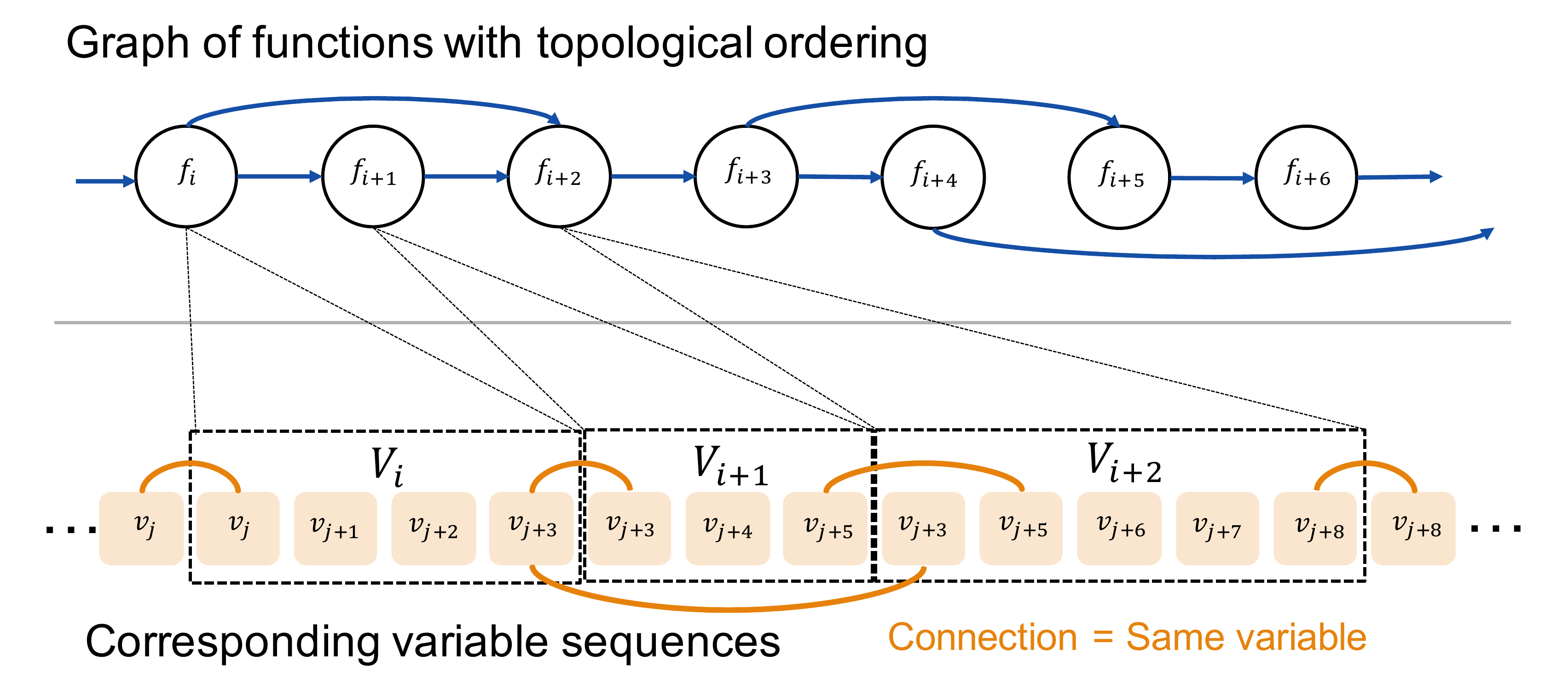}
\caption{(Top) Directed graph of functions in topological order. Any feed-forward neural networks can be considered as a sequence of functions and can be executed from $f_1$ to $f_n$ in order. (Bottom) A sequence of variables corresponding to the sequence of functions. Each function uses variable internally and different functions can have the same variable (connection between variables). We estimate distances between functions by sum of bytes of variables within two functions (e.g. we estimate the distance between $f_i$ and $f_{i+3}$ as the sum of bytes of $\{V_{i+1}, V_{i+2}\}$ in this picture).}
\label{fig:graph}
\end{figure}

\section{Preliminaries}

Feed-forward neural networks can in general be represented as a directed acyclic graph (DAG) G(V, E) of functions (Fig.\ref{fig:graph} (Top)), where V is a set of function $\{f_i\}_{i = 1}^{N}$ and there is an edge from $f_i$ to $f_j$ if $f_j$ uses an output of $f_i$ as an input. Through topological ordering, we can index all functions in ascending order from inputs to outputs of entire network such that if there is an edge from $i$ to $j$, then we always have $i < j$. Thus, we can simply consider a given network as a sequence of functions $[f_1, \dots, f_n]$ and can train the network by executing functions from $f_1$ to $f_n$ in order. Out-of-core execution on DNNs can utilize this fact of function order as a strong assumption to find a better schedule for memory swapping. Namely, we can select which memory we should transfer from GPU to CPU (\emph{Swap-out}) and decide when we should start transferring a memory from CPU to GPU (\emph{Swap-in}) on the given topological order of functions.

Some early works focus on forward and backward process of DNNs \citep{Rhu2016vdnn,wang2018superneuron,Jin2018LayRub}. They employ a simple memory swap scheduling where they perform \emph{Swap-out} for the outputs of forward functions and \emph{Swap-in} at corresponding backward functions. While these methods enable the training of larger models than the setting without out-of-core execution, there are two drawbacks on these methods. First, when a variable is used multiple times in a forward pass, they perform \emph{Swap-out} only at the last usage, which means the data of the variable is kept on GPU memory until the last usage during the forward computation (even for backward). This is problematic when the network is very large and it has, for example, skip connections which are seen in currently popular models such as DenseNet \citep{huang2017densely} and U-Net like architectures \citep{MICCAI15UNet}. Second, they do not carefully handle the overhead caused by \emph{Swap-in}. In order to make data for a variable ready on GPU memory before starting computation, they use a naive heuristic where they trigger a \emph{Swap-in} operation for a variable used in a function right before the previous function of that (e.g. trigger \emph{Swap-in} for $f_i$ at $f_{i-1}$). It causes large overhead if $f_{i-1}$'s computation is too small compared to the memory transfer latency for \emph{Swap-in}, i.e. the function has to wait for the completion of the memory transfer.

\citet{IBM2019} recently proposed \emph{Large Model Support} (LMS) as a module on TensorFlow, in which they consider a distance on a graph (DAG) of a neural network to determine the memory swap scheduling which handles the two issues described above. They introduce two hyperparameters: \emph{swapout\_threshold} and \emph{swapin\_ahead}. They selectively trigger \emph{Swap-out} on a variable used at a function $f_i$ when $d(f_i, f_j) \leq $ \emph{swapout\_threshold} where $d(\cdot, \cdot)$ is a distance on the graph between two vertices and $f_j$ is a closest function which uses the variable used in $f_i$ ($i < j$). This allows triggering \emph{Swap-out} at any point of the entire graph computation including forward and backward, and enables training larger models with skip connections. Then, they trigger \emph{Swap-in} for the swapped out variable right before executing a function $f_k$ where $d(f_k, f_j) = $ \emph{swapin\_ahead} and $k < j$. Consequently, they successfully train ResNet50 with 4.9 times larger batch-size and with less overhead, which has not been achieved by the previous methods.

While \cite{IBM2019} achieves promising results, there still remain the following two issues: 1) They rely on a fixed distance across the entire graph to determine the memory swap scheduling, while an actual cost (latency) for computation and memory transfer between two vertices varies depending on operations and data sizes appeared between them, which may lead undesirable overhead due to the gap between the cost of computation and memory transfers. 2) Both of two hyperparameters mentioned above balance a trade-off between reducing memory transfer overhead and GPU memory usage. Increasing either of the hyperparameters may result in out-of-memory error. Finding good hyperparameters with less overhead and without raising memory error is difficult for humans. Therefore they introduce an automatic tuning mechanism which relies on a fairly complicated simulation based on memory profiling and computational cost estimation for each operator manually defined by humans.




\section{Scheduling memory swapping based on locally-adaptive window}
\label{Scheduling}

To overcome the aforementioned issues, we propose a novel memory transfer scheduling algorithm which adaptively determines distance thresholds for the \emph{Swap-in} and \emph{Swap-out} operations while considering memory budget limitation.
Aside from a sequence of functions $[f_1, \dots, f_N]$ which we call \emph{function-sequence} later, we introduce a sequence of all variables used in order in the \emph{function-sequence}, which we call \emph{variable-sequence}.
Let $ V = \{v_j\}_{j = 1}^{N_v}$ be a set of all variables used on a given network and  $\hat{V}_i \subseteq V$ be a subset of variables used by $f_i$. Given a topological order of functions $[f_1, \dots, f_n]$, we can also define a topological sequence of variables (i.e., \emph{variable-sequence}) $\boldsymbol{v} = flatten([\hat{V_1}, \dots, \hat{V_n}])$ where $flatten()$ represents flatten function (representing given matrix as 1-dimensional array). Since multiple functions can use the same variable either as input or output, the same variable can appear multiple times in this sequence (Fig.\ref{fig:graph} (Bottom)). Each variable has its size in bytes, and a state representing whether it is on GPU or CPU now. We denote the size in bytes of variable $v_j$ as $b_{v_j}$, and state of variable $\sigma(v_j) \in \{0, 1\}$, which returns $1$ if  $v_j$ is placed on CPU, otherwise $0$.

Given a \emph{variable-sequence}, our goal is to find a better schedule for which variables we should apply \emph{Swap-in} and \emph{Swap-out} at each function. Unfortunately, finding the optimal schedule for \emph{Swap-in} and \emph{Swap-out} on $\boldsymbol{v}$ is difficult, due to the huge search space. Let $V_i^{'}$ be a set of variables which never appears from function $f_i$, and $n$ be the number of functions. At each function, we must consider which variables are on GPU. Therefore, the entire search space can be represented as  $O(\prod_{i=1}^{n}{2^{|V \setminus V_i^{'}|}})$. This is not tractable, since modern neural networks have more than a hundred functions and variables.

\begin{figure}[t!]

\begin{subfigure}{0.5\textwidth}
    \includegraphics[width=0.9\linewidth]{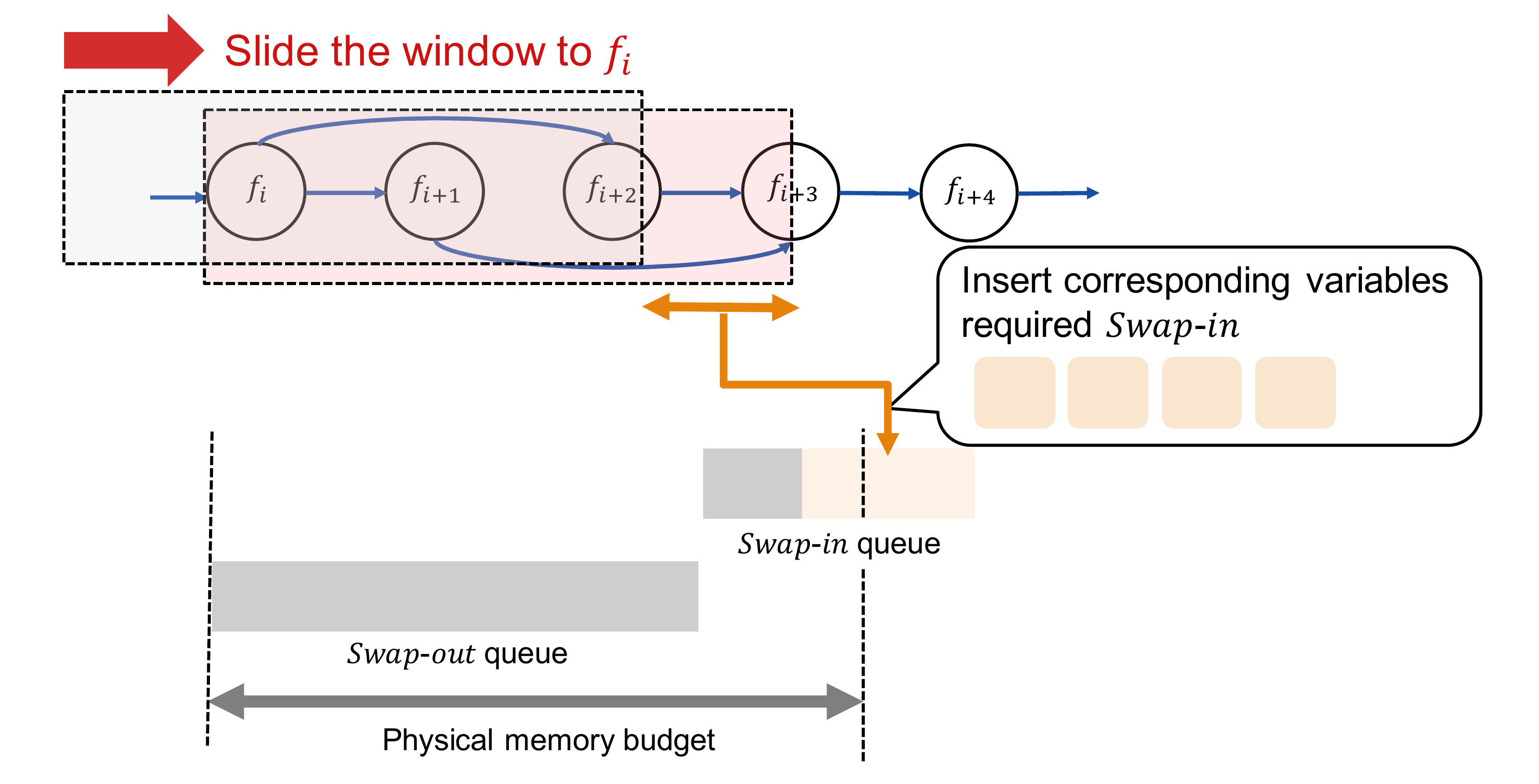} 
    \caption{Schedule \emph{Swap-in}} 
    \label{fig:subim1}
\end{subfigure}
\begin{subfigure}{0.5\textwidth}
    \includegraphics[width=0.9\linewidth]{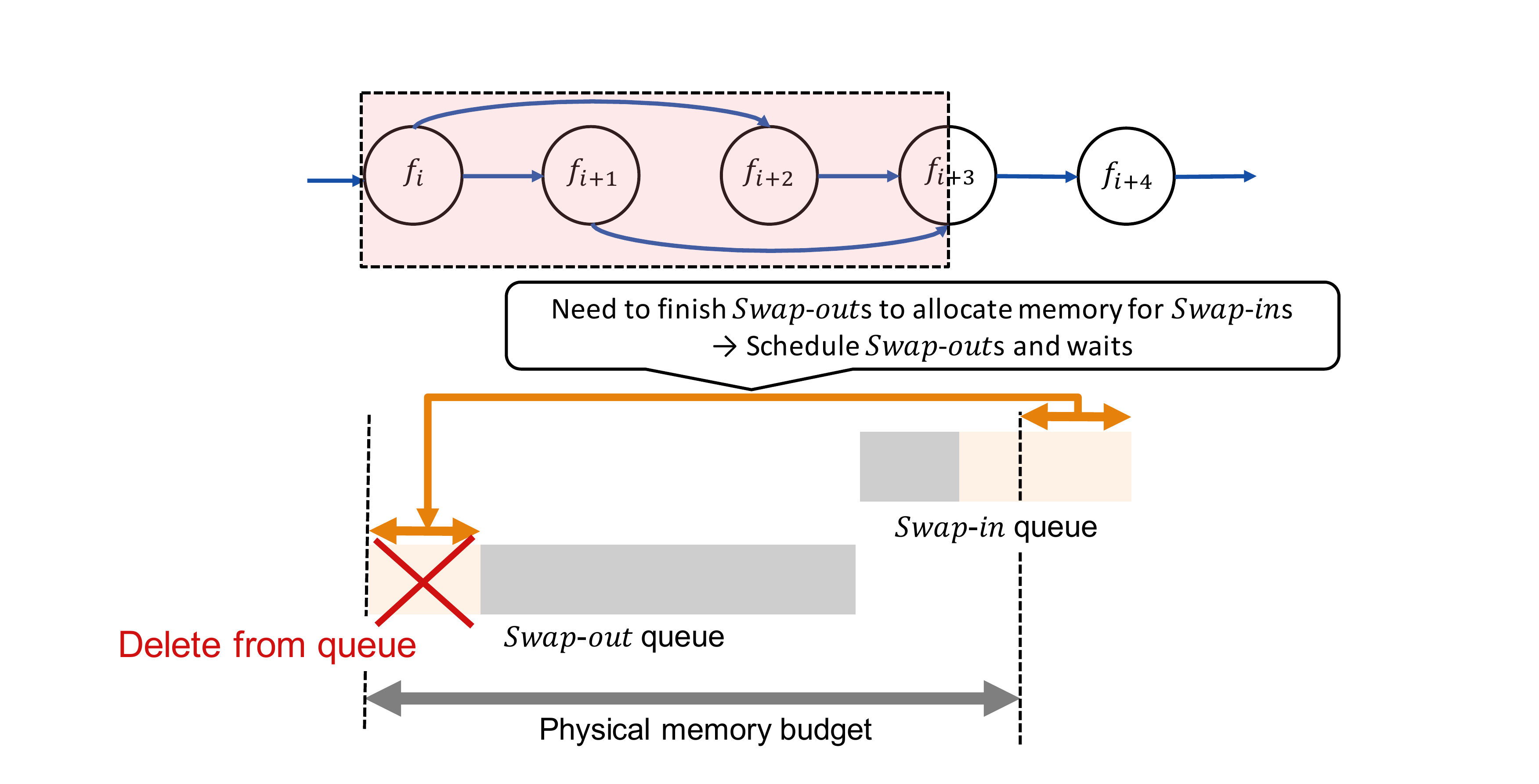}
    \caption{Determine \emph{Swap-out}}
    \label{fig:subim2}
\end{subfigure}
\begin{subfigure}{0.5\textwidth}
    \includegraphics[width=0.9\linewidth]{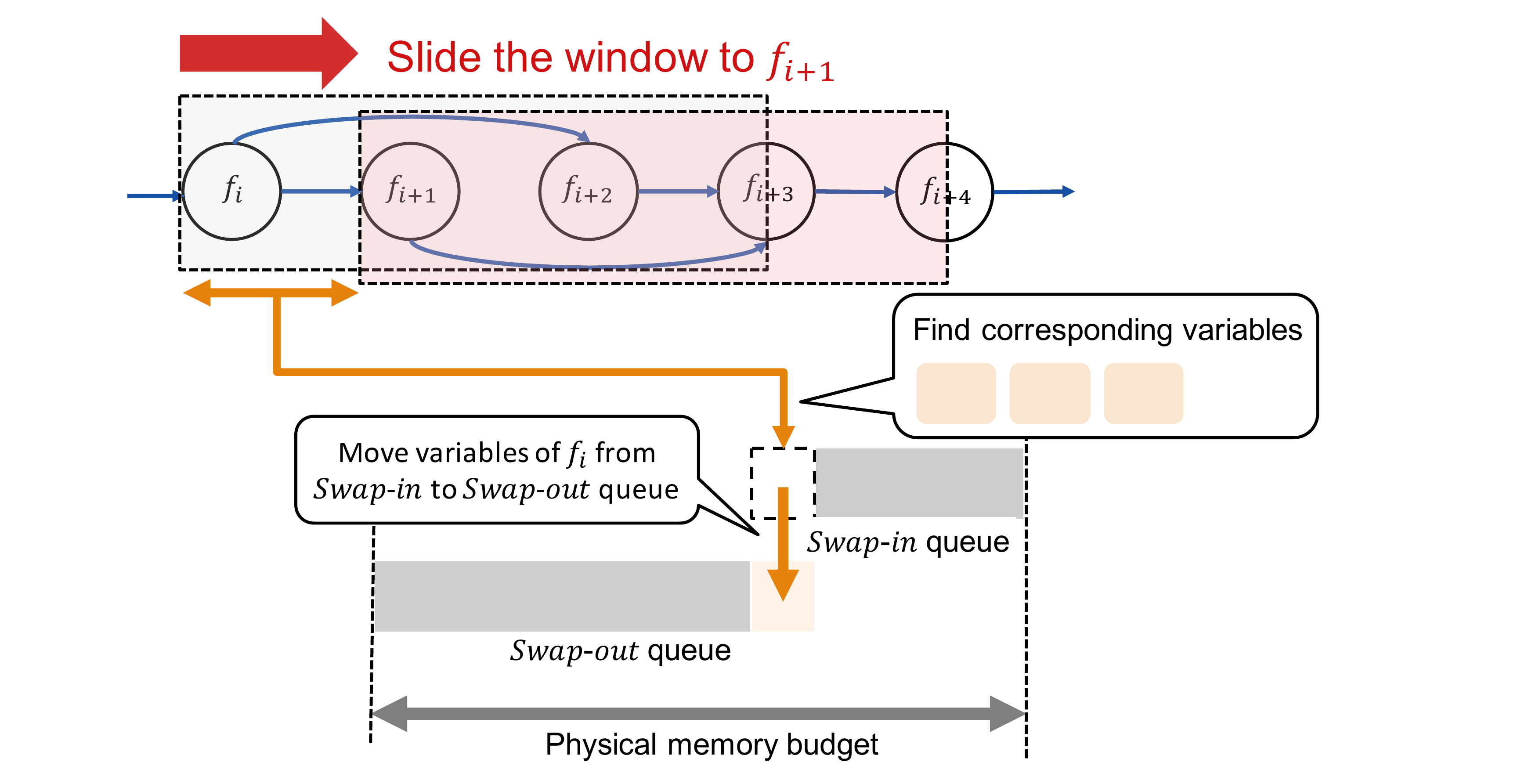}
    \caption{Schedule \emph{Swap-out}}
    \label{fig:subim3}
\end{subfigure}
\begin{subfigure}{0.5\textwidth}
    \includegraphics[width=0.9\linewidth]{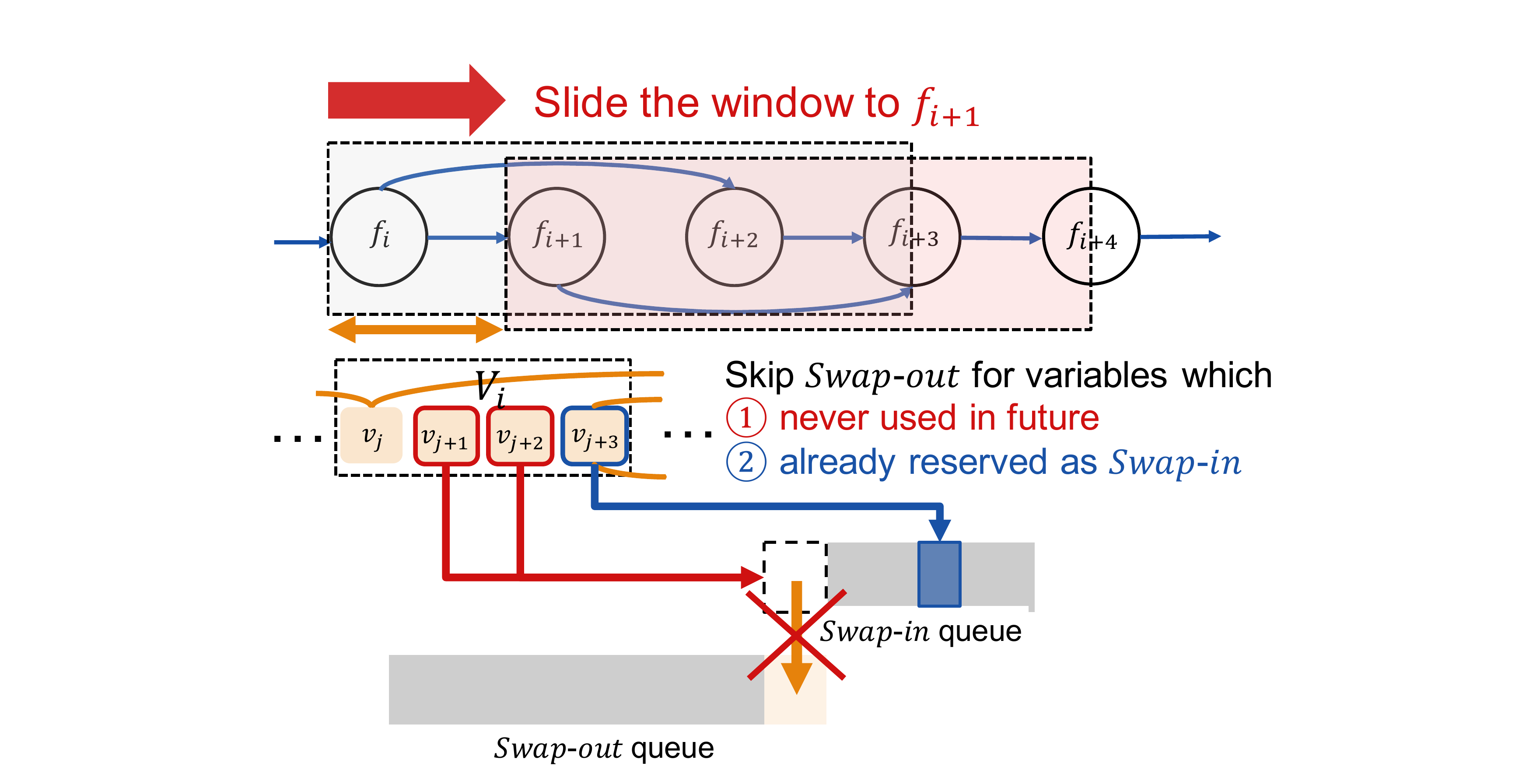}
    \caption{Skip scheduling}
    \label{fig:subim4}
\end{subfigure}
\caption{An overview of our scheduling algorithm at function $f_i$. Red rectangle shows \emph{schedule-window}. Sliding this window from $f_1$ to $f_n$, our algorithm performs (a) $\rightarrow$ (b) $\rightarrow$ (c) at each function to schedule \emph{Swap-in} and \emph{Swap-out}. (a) Schedule \emph{Swap-in} for all new variables coming into the window. (b) If scheduling exceeds physical memory budget by scheduling \emph{Swap-in}, we allocate enough memory for \emph{Swap-in} by determining \emph{Swap-out} for previous variables. Specifically, \emph{Swap-out} is reserved right after the previous function using the variable, and waits for the \emph{Swap-out} right before $f_i$. (c) Schedule \emph{Swap-out} for variables used by $f_i$ and move the window to the first variable of next function. (d) shows how we reduce unnecessary memory transfers. We skip \emph{Swap-out} for the variables used by $f_i$ if variables are never used in future or already reserved as \emph{Swap-in}.}
\label{fig:schedule}

\end{figure}

In this paper, we employ a simple local greedy search in \emph{variable-sequence} which requires only a single hyperparameter \emph{schedule-window}. At first, let us focus on when to trigger \emph{Swap-in} for variables swapped out previously. It is most ideal that we trigger \emph{Swap-in} such that the computation starts immediately after the completion of \emph{Swap-in} memory transfer. If we trigger \emph{Swap-in} too early, it unnecessarily consumes more memory, while more overhead if too late. In order to determine this ideal timing, we have to precisely estimate computation time and memory transfer time, which is non-trivial. Instead, we use a very simple assumption where both computation and memory transfer time are roughly proportional to the data size involved with them. We trigger \emph{Swap-in} operations at a function $f_i$ by looking-ahead \emph{variable-sequence} until accumulated variable size in bytes exceeds the threshold \emph{schedule-window}. More formally, we trigger \emph{Swap-in} for variables in a sub-sequence $\boldsymbol{v}[l:r]$ where $l$ is the index of the first variable of $\hat{V}_i$ on $\boldsymbol{v}$, and $r$ is the maximum index which satisfies $\sum_{i = l}^{r} b_{\boldsymbol{v_i}} \leq$ \emph{schedule-window}.

Fig.\ref{fig:schedule} shows an overview of our scheduling procedure at $f_i$. Let the sub-sequences determined by \emph{schedule-window} at $f_{i-1}$, $f_{i}$, and $f_{i+1}$ be $\boldsymbol{v}[l^-:r^-]$, $\boldsymbol{v}[l:r]$, and $\boldsymbol{v}[l^+:r^+]$, respectively. Our scheduling procedure at $f_i$ comprises following three steps: First, we schedule \emph{Swap-in} for variables $\{v_{si} \in \boldsymbol{v}[r^-+1:r] \ |\ \sigma(v_{si}) = 1\}$ (Fig.\ref{fig:subim1}). Second, we determine the completion of previously scheduled \emph{Swap-out} (oldest) such that the total variable size of \emph{Swap-in} and \emph{Swap-out} doesn't exceed the physical GPU memory budget (Fig.\ref{fig:subim2}). These memory transfers are scheduled to be performed before executing $f_i$. Finally, we reserve \emph{Swap-out} for $\hat{V_i}$, which is the same as $\{v_{so} \in \boldsymbol{v}[l:l^+-1]\}$, after executing $f_{i}$ and move \emph{schedule-window} to the next function $f_{i+1}$ (Fig.\ref{fig:subim3}). To reduce redundant memory transfers, we skip schedules for variables if the same variables are already scheduled (Fig.\ref{fig:subim4}).

Compared to \cite{IBM2019}, our algorithm determines both \emph{Swap-in} and \emph{Swap-out} scheduling according to only a single hyperparameter with a physical memory budget constraint. Also, note that our algorithm adaptively controls the \emph{Swap-in} distance threshold at each function in \emph{function-sequence} in order to consider computation and memory transfer time for efficient scheduling, while we simply use a fixed-size window in bytes in \emph{variable-sequence}. 

\section{Reducing memory fragmentation with virtual addressing}
\label{VirtualAddressing}

Most deep learning frameworks (e.g. Tensorflow \citep{tensorflow2015-whitepaper}, PyTorch \citep{NIPS2019Pytorch}, and Neural Network Libraries) commonly utilize the ``best-fit" algorithm with caching GPU memory as a memory allocation strategy. This approach utilizes memory space effectively by reusing previously allocated memory for multiple variables. It performs well when the number of reusing the same memory space is limited. However, in the setting with out-of-core execution, frequent memory transfers between CPU and GPU result in reusing the same memory beyond the number of times manageable by the best-fit algorithm. Thus, the memory allocation system will divide its cached memory repeatedly, which results in severe memory fragmentation. This fragmentation is generally known as External Fragmentation in the field of OS memory allocation. It is also important to tackle this memory fragmentation problem to maximize trainable model size under fixed memory budget.

Since estimating exact amount of external fragmentation is difficult, we estimate it as the worst case. Let a requested size of bytes be $m_r$. In the worst case, when this $m_r$ cannot be used for any successive variables and also cannot be merged with consecutive memories, this $m_r$ memory is kept in cache without being used during the training. Thus, the memory request of $m_r$ bythe would waste $m_r$ bytes in the worst case. When we grow model size with out-of-core execution, $m_r$ could be in order of GBs.

To tackle this external memory fragmentation, we apply virtual addressing (VA), which is commonly used in memory management on OS, to the training of DNNs. Namely, we manage both physical and virtual memory addresses on neural network libraries. When the memory allocation is requested, we map small physical memory chunks with constant size to a consecutive virtual address. Once a variable is cleared during the training of neural network, we release a virtual address and cache physical memories for the future requests. Since physical memories could be combined in arbitrary order, virtual addressing never suffers from external fragmentation on physical memory address. In consequence, we can simply estimate the amount of fragmentation in virtual addressing by the difference between requested memory size and the size of allocated virtual address for this request (this memory fragmentation is known as internal fragmentation in OS).

Let $m_c$ be a size of physical memory chunk. To minimize the amount of wasted memory by internal fragmentation, we allocate virtual address for the request as $m_a = km_c$ where $k = \lceil \frac{m_r}{m_c} \rceil$. In this case, we can estimate the amount of internal fragmentation ($IF$) for a single allocation request as:

\begin{eqnarray}
  IF &=& m_a - m_r < m_c.
\end{eqnarray}

It is notable that the upper-bound of $IF$ doesn't depend on $m_r$. Hence, the maximum size of internal fragmentation during training ($IF_{max}$) can be bounded by

\begin{eqnarray}
  IF_{max} &<& N_{max}m_c,
\end{eqnarray}

where $N_{max}$ is the maximum number of variables that can be used on GPU memory simultaneously. In general, this is suitable for the training of extremely large-scale neural networks under out-of-core execution. In the out-of-core execution, we perform \emph{swap-out} and only variables within \emph{Schedule-window} are placed on GPU. Therefore, $N_{max}$ becomes smaller as model size becomes larger, and $IF_{max}$ also approaches a small value.

Obviously, it is better to use smaller physical memory chunk to reduce internal fragmentation. However, there is a restriction on device. On CUDA Driver API, the minimum size of physical memory chunk is defined depending on the type of GPU. In our environment, the minimum size of physical memory chunk is 2MB. Besides, using smaller physical memory chunks causes additional overhead for virtual address mapping. We experimentally decide the size of physical memory as 40MB, which balances well between the cost of allocation and the amount of memory fragmentation.

\section{Experiments}
All experiments were conducted on an IBM POWER9 machine with 594GB of CPU RAM and NVIDIA Tesla V100 GPUs (each GPU has 16GB of memory). Note that we only used single GPU for all experiments. CPU and GPU are connected by two NVLinks (each can transfer memory at 50GB/s). We employ Neural Network Libraries\footnote{https://github.com/sony/nnabla} (NNL) as our deep learning framework under CUDA Toolkit 10.2 and cuDNN v8.0.2.

In all experiments, we evaluate both scheduling alone and scheduling with virtual addressing described in Section \ref{VirtualAddressing}. To evaluate scheduling alone, we use the default memory allocator on NNL, which applies ``best-fit" algorithm with caching memory in nearly identical manner as other deep learning frameworks, such as Torch or TensorFlow.

\begin{table}
\begin{center}
\begin{tabular}[t!]{ |c||c|c|c|c|c|c|c|c|c|c| } 
 \hline
 Batch size & 64 & 128 & 190 & 256 & 512 & 928 & 1120 & 1248 & 1440 \\ \hline \hline
 Baseline & 235 & 255 & 258 & x & x & x & x & x & x \\
 LMS \citep{IBM2019} & 235 & 255 & 241 & 236 & 209 & 137 & x & x & x \\  
 \hline
 Baseline & 336 & 492 & 581 & x & x & x & x & x & x \\ 
 Schedule & 320 & 439 & 533 & 559 & 457 & 400 & x & x & x \\
 Schedule + VA & 304 & 389 & 415 & 424 & 446 & 409 & 398 & 347 & 321 \\
 \hline
\end{tabular}
\caption{Training time and model size comparison between Le's method and ours. We use images / sec (ips) for all values on the table, where x indicates Out-of-Memory and thus cannot be trained. The top 2 rows show Le's results and bottom 3 rows show ours. Since computational environments are different (P100 vs V100), we show the results of baseline for both methods separately. All results for \cite{IBM2019} are from the best values reported on the paper.}
\label{table: resnet50}
\end{center}
\end{table}

\begin{figure}[t]
\begin{subfigure}{0.5\textwidth}
\includegraphics[width=0.9\linewidth]{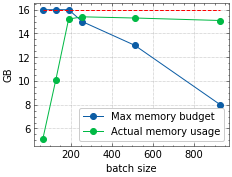}
\caption{Maximum defined memory budget and actual memory usage in the setting with scheduling alone.}
\label{fig:resnet_ca}
\end{subfigure}
\begin{subfigure}{0.5\textwidth}
\includegraphics[width=0.9\linewidth]{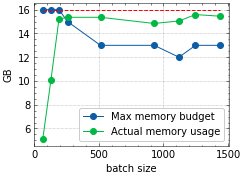}
\caption{Maximum defined memory budget and actual memory usage in the setting with scheduling with VA.}
\label{fig:resnet_va}
\end{subfigure}
\caption{Maximum defined memory budget and actual memory usage. In the setting with scheduling alone, there are almost 2$\times$ differences at 928 batch size. This indicates memory fragmentation. On the other hand, scheduling with VA prevents such fragmentation.}
\label{fig:exp:res_mem}
\end{figure}

\subsection{Imagenet classification}
We first compare our method with Le's algorithm on ResNet-50 training on ImageNet \citep{imagenet_cvpr09}. We keep image size as $224\times224$ and increase only batch size, which is the same condition reported by \cite{IBM2019}. Both methods are examined on the same GPU memory budget (16GB). We examined our method on NVIDIA Tesla V100, while \cite{IBM2019} uses P100. Thus, all computations are theoretically faster on our environment. This however implies a more challenging criterion for our model, as will be described below.

Table.\ref{table: resnet50} shows comparison between Le's algorithm and ours in terms of training time and trainable model size. We first validate our scheduling without VA. Our scheduling not only successfully trains the maximum size of model which Le's approach can train (batch size = 928), but also finds a faster schedule than Le's approach. Our schedule can keep training speed at 68\% at 928 batch size, while Le's method performs at 53\%. It is noteworthy that comparing results with absolute value of ips is also reasonable, because in out-of-core execution, our primary interest is how much memory transfers we can overlap with computation. Since the model size and memory budget are the same on both methods, faster computational environment means that we have to transfer the same amount of memory within shorter time. Even with such disadvantage, our schedule clearly outperforms Le's method in terms of absolute ips.

While our method successfully finds faster schedule, scheduling alone cannot train larger model than 928 batch size as well as Le's method.
Fig.\ref{fig:exp:res_mem} shows maximum memory budget we can define for scheduling (blue line) and actual memory usage (green line). Red line shows physical memory budget. Fig.\ref{fig:resnet_ca} shows the results in the setting with scheduling alone. At 928 batch size, network consumes almost 16GB while we set 8GB as a memory budget for scheduling. In fact, if we set larger than 8GB for 928 batch size, training process causes Out-of-Memory even though a schedule is found. This difference between the memory budget we set and actual usage indicates the memory fragmentation and this is the reason why we cannot train a model with batch size lager than 928.
On the other hand, using our novel VA allocator with scheduling, we successfully train the same model with 1,444 batch size, which is 1.6$\times$ larger than previous limitation with small overhead. Moreover, our scheduling with VA enables faster training at batch size of 928 against scheduling alone. This is because we can set much more physical memory budget for scheduling. As we can see on Fig.\ref{fig:resnet_va}, the difference between the memory budget we can set for scheduling and actual usage is smaller, even when we increase batch size. This indicates that VA drastically reduces fragmentation and we can conclude that VA is beneficial for out-of-core execution, especially for training extremely large-scale DNNs. Note that if we employ smaller $m_c$, this difference becomes smaller and we can train much larger models, but the overhead in terms of training time becomes larger due to the cost of virtual addressing.  

\subsection{Training time and model size on various neural network architectures}

\begin{figure}[t]
\begin{subfigure}{0.33\textwidth}
\includegraphics[width=0.9\linewidth]{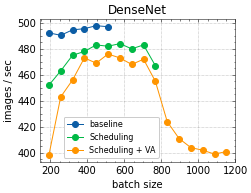}
\label{fig:densenet_speed}
\end{subfigure}
\begin{subfigure}{0.33\textwidth}
\includegraphics[width=0.9\linewidth]{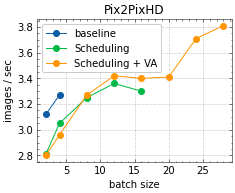}
\label{fig:pix2pixHD_speed}
\end{subfigure}
\begin{subfigure}{0.33\textwidth}
\includegraphics[width=0.9\linewidth]{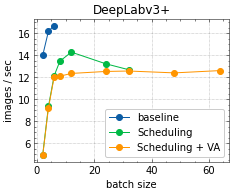}
\label{fig:deeplab_speed}
\end{subfigure}
\caption{Training time as model size grows for various architectures. We evaluate average iteration time including forward, backward, and update over 100 iterations.}
\label{fig:exp:speed}
\end{figure}

\begin{figure}[t]
\begin{subfigure}{0.33\textwidth}
\includegraphics[width=0.95\linewidth]{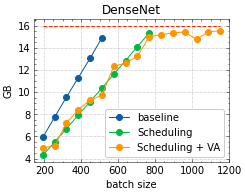}
\label{fig:DenseNet_mem}
\end{subfigure}
\begin{subfigure}{0.33\textwidth}
\includegraphics[width=0.9\linewidth]{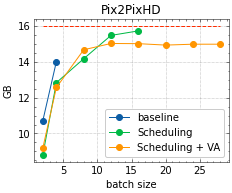}
\label{fig:Pix2PixHD_mem}
\end{subfigure}
\begin{subfigure}{0.33\textwidth}
\includegraphics[width=0.9\linewidth]{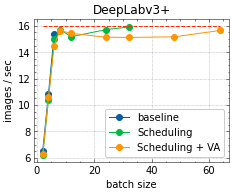}
\label{fig:deeplab_mem}
\end{subfigure}
\caption{Memory usage as model size grows for various architectures. We evaluate the peak memory usage during forward, backward, and update for each network. Red dashed line represents physical memory budget (16GB).}
\label{fig:exp:memory}
\end{figure}

We examined our proposed method over various network architectures in terms of training time and memory usage as the model size grows. Followings are the list of network architectures we examined: DenseNet \citep{huang2017densely}, Pix2PixHD \citep{wang2018pix2pixHD}, and DeepLabv3+ \citep{chen2018encoder}. For all networks, we increased batch size with fixed image size and all computations are executed with float32 precision. As training dataset for each model, we use ImageNet with $224\times224$ as image size for DenseNet, Cityscapes dataset \citep{Cordts2016Cityscapes} with $512\times1024$ for Pix2PixHD, and PASCAL VOC dataset \citep{pascalvoc2010} with $513\times513$ for DeepLabv3+.

Fig.\ref{fig:exp:speed} shows training time against batch-size on 3 models. The horizontal axis represents batch size, and the vertical axis represents ips for a single training step, including forward, backward and update. In the baseline setting (blue line), the maximum batch sizes trained on 16GB memory for DenseNet, Pix2PixHD, and DeepLabv3+ are 512, 4, and 6, respectively. With our scheduling (green line), we can successfully train the models that are 1.5x, 4x, and 5.3x larger than baseline setting. However, as we can see in Fig.\ref{fig:exp:memory}, training with scheduling alone reaches the physical memory limit because of memory fragmentation. Applying proposed VA with our scheduling (orange line), trainable batch size clearly improves with small overhead for all networks. Compared to the setting with scheduling alone, our scheduling with VA can train 1.5x to 2x larger models. Fig.\ref{fig:exp:memory} shows that we can keep the memory usage almost constant  with VA regardless of batch size. It is notable that in pix2pixHD we achieve performance gain compared to baseline setting. We consider that larger batch size is computationally beneficial, especially for slower networks (Pix2PixHD achieves only around 3 images / sec).

\section{Conclusion}
In this paper, We propose a novel out-of-core algorithm that enables faster training of extremely large-scale neural networks with sizes larger than allotted GPU memory. Under a given memory budget constraint, our scheduling algorithm locally adapts the timing of memory transfers according to memory usage of each function, which improves overlap between computation and memory transfers. 
Additionally, we apply virtual addressing technique, commonly performed in OS, to training of neural networks with out-of-core execution, which drastically reduces the amount of memory fragmentation caused by frequent memory transfers. Beyond GPU memory limitation, we empirically show that our proposed method enables training of much larger networks than existing methods, without sacrificing the training time. While our proposed algorithm clearly demonstrates improvements over previous models, it is still not optimal, e.g., we ignore the order of variables for \emph{Swap-out}. We leave further optimization as future work.



\bibliography{lms}
\bibliographystyle{iclr2021_conference}

\end{document}